\documentclass[10pt,twocolumn,letterpaper]{article}

\usepackage[pagenumbers]{cvpr} 

\definecolor{cvprblue}{rgb}{0.21,0.49,0.74}
\usepackage[pagebackref,breaklinks,colorlinks,allcolors=cvprblue]{hyperref}

\usepackage{color}
\usepackage{xcolor}
\usepackage{colortbl}
\definecolor{Gray}{gray}{0.9}
\usepackage{threeparttable}
\usepackage{multirow}
\usepackage{array}
\usepackage{makecell}
\newcolumntype{C}[1]{>{\centering}p{#1}}
\usepackage{epsfig}
\usepackage{graphicx}
\usepackage{float}
\usepackage{tcolorbox}
\usepackage{appendix}
\definecolor{ours}{RGB}{244,237,252}
\definecolor{image-level}{RGB}{247, 247, 247}
\definecolor{region-level}{RGB}{239, 245, 234}

\def\paperID{6859} 
\def\confName{CVPR}
\def\confYear{2025}

\title{LLMDet: Learning Strong Open-Vocabulary Object Detectors under the Supervision of Large Language Models}

\author{%
Shenghao Fu$^{1,2,4}$, Qize Yang$^{2}$, Qijie Mo$^{1,4}$, Junkai Yan$^{1,4}$, Xihan Wei$^{2}$, \\Jingke Meng$^{1,4}$,
{Xiaohua Xie$^{1,4,5}$\thanks{: Corresponding authors are Xiaohua Xie and Wei-Shi Zheng.\\\indent Part of the work was done when Shenghao Fu was an intern at Alibaba. }, Wei-Shi Zheng$^{1,3,4,6*}$} \\
$^1$School of Computer Science and Engineering, Sun Yat-sen University, China;\\
$^2$Tongyi Lab, Alibaba Group; \quad $^3$Peng Cheng Laboratory, China;\\
$^4$Key Laboratory of Machine Intelligence and Advanced Computing, Ministry of Education, China;\\
$^5$Guangdong Province Key Laboratory of Information Security Technology, China; \\
$^6$Pazhou Laboratory (Huangpu), China\\
fushh7@mail2.sysu.edu.cn, xiexiaoh6@mail.sysu.edu.cn, wszheng@ieee.org
}

\begin{document}
\maketitle

\begin{abstract}
Recent open-vocabulary detectors achieve promising performance with abundant region-level annotated data. 
In this work, we show that an open-vocabulary detector co-training with a large language model by generating image-level detailed captions for each image can further improve performance. To achieve the goal, we first collect a dataset, GroundingCap-1M, wherein each image is accompanied by associated grounding labels and an image-level detailed caption. With this dataset, we finetune an open-vocabulary detector with training objectives including a standard grounding loss and a caption generation loss. 
We take advantage of a large language model to generate both region-level short captions for each region of interest and image-level long captions for the whole image. Under the supervision of the large language model, the resulting detector, LLMDet, outperforms the baseline by a clear margin, enjoying superior open-vocabulary ability. Further, we show that the improved LLMDet can in turn build a stronger large multi-modal model, achieving mutual benefits. The code, model, and dataset is available at \url{https://github.com/iSEE-Laboratory/LLMDet}.
\end{abstract}

\section{Introduction}

\begin{figure}[t]
  \centering
  \includegraphics[width=1\linewidth]{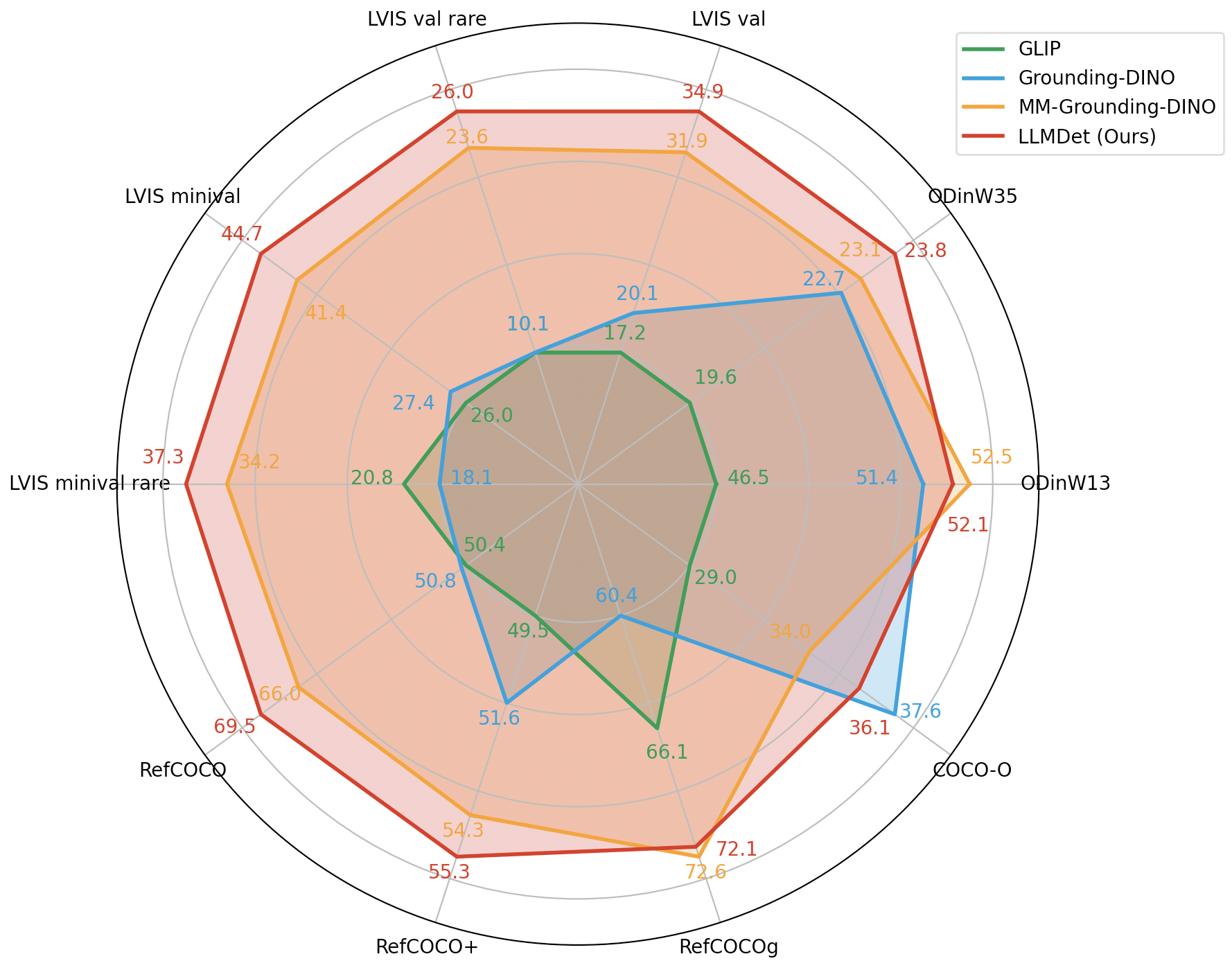}
  \caption{LLMDet achieves superior zero-shot performance across various benchmarks compared with other well-known counterparts. All detectors use Swin-T as the backbone.}
  \vspace{-0.6em}
  \label{fig:compare_performance}
\end{figure}

Open-vocabulary object detection~\cite{vild, GLIP, grounding_dino} aims to detect arbitrary classes based on text labels from user input, which is a more general detection task than traditional closed-set object detection~\cite{he2017mask, ross2017focal, detr, fu2023asag}. GLIP~\cite{GLIP} first unifies object detection and phrase grounding through region-word contrastive pre-training. This formulation benefits from massive grounding and image-text data covering a broad range of concepts, making the learned representations semantic-rich. The following works focus on effective vision-language fusion~\cite{fiber, grounding_dino} and fine-grained region-word alignment by carefully designed word embeddings~\cite{yao2022detclip} and negative samples~\cite{li2024desco, yao2023detclipv2, zhao2024generating}. By scaling up pretraining data and computation~\cite{yao2023detclipv2, OWL-ST}, existing open-vocabulary object detectors can achieve amazing zero-shot performance on various benchmarks.

Recent works show that unifying the grounding task with other language tasks enriches visual representations with language knowledge thus creating a stronger open-vocabulary detector. GLIPv2~\cite{zhang2022glipv2} pretrains the model under the grounding loss and the masked language modeling loss. Subsequently, CapDet~\cite{long2023capdet} and DetCLIPv3~\cite{yao2024detclipv3} demonstrate that unifying dense captioning and grounding also boosts open-vocabulary ability. 
However, they use short captions for each object, \eg, coarse descriptions and hierarchical class labels, which are coarse-grained, individual, and lacking association among objects. Alternatively, long image-level captions, containing rich details and a comprehensive understanding of an image, provide more information than short region-level descriptions, which motivates us to explore what advantages can long detailed image-level captions bring to open-vocabulary detectors.

In light of this, we propose LLMDet, which trains an open-vocabulary detector with a standard grounding objective accompanied by a \textbf{caption generation objective}.
A large language model (LLM) is appended to the detector, takes both image features and region features from the detector as input, and predicts image-level long detailed captions and region-level short phrases, separately.
Compared with previous works that only generate short captions for each object, our LLMDet excels from four aspects: \textbf{First, long captions provide more details for each object in the image.} Long captions with detailed \textit{object types, textures, colors, parts of the objects, object actions, precise object locations, and texts} are helpful to build rich vision-language representations. While existing region-level captions are overly simplistic descriptions for regions. \textbf{Second, image-level generation aligns all elements in the image as a whole}, which models both foreground objects, background and the relations between various objects, providing more information and a more comprehensive understanding of the image beyond object-level caption that only focuses on single regions of interest. \textbf{Third, image-level captions are more scalable than region-level annotations.} Recent off-the-shelf large vision-language models excel at whole image understanding but still struggle with precise region-level understanding. With proper prompts, we can get high-quality image-level captions from them at a low cost. \textbf{Further, the fully-pretrained large language model is naturally open-vocabulary.} Using an LLM to generate captions makes the detector align with it, thus inheriting a strong generalization ability and significantly increasing rare class performance.

However, existing grounding datasets lack detailed captions for the whole image. Thus, we first collect a dataset, named GroundingCap-1M, to train LLMDet. Compared with a standard grounding dataset, each element in GroundingCap-1M is formulated as a quadruple, containing an image, a short grounding text, some annotated bounding boxes mapped to the phrases in the grounding text and a long detailed image-level caption. A large language model is utilized to understand region and image features and generate the grounding phrases corresponding to each object and the image-level caption. To effectively integrate the large language model into LLMDet and preserve the pretrained knowledge, we first carefully align the large language model with an existing detector and then finetune them as a whole.

With the novel training framework, \textbf{we show that the vision foundation model can benefit from the supervision of LLMs.} The supervision not only comes from using LLM-generated captions as labels but also comes from the gradient of co-training. The resulting LLMDet achieves outstanding open-vocabulary performance. Compared with the baseline, LLMDet outperforms 3.3\%/3.8\%/14.3\% AP and 3.1\%/3.3\%/17.0\%  AP$_r$ with Swin-T/B/L as backbone on LVIS~\cite{gupta2019lvis} minival. We also perform a comprehensive zero-shot transfer to various datasets to demonstrate LLMDet's superior performance, as shown in \Cref{fig:compare_performance}.

By integrating the improved LLMDet with a large language model, we can further build a strong large multi-modal model (LMM). Training under the supervision of large language models, LLMDet not only achieves stronger open-vocabulary ability but also pre-aligns with the large language models. Thus the pretrained LLMDet can serve as a strong vision foundation model and in turn build a better LMM, achieving mutual benefits (shown in Appendix).

\section{Related Work}

\subsection{Open-Vocabulary Object Detection}
In open-vocabulary object detection (OVD), the detector is trained on a limited training dataset but aims to detect arbitrary test-time user-input classes. To detect arbitrary classes, open-vocabulary object detection is formulated as a vision-language task so that the detector can detect classes never seen with class names. Motivated by the impressive zero-shot ability of vision-language models (\eg CLIP~\cite{clip}), aligning detectors with CLIP~\cite{vild, BARON, dk-detr} or integrating CLIP as a part of model~\cite{fvlm, fu2024frozen-detr} are straight-forward directions for addressing OVD. However, since CLIP is pretrained with image-level objectives, the features in CLIP are not perfectly suitable for OVD. 

Alternatively, building an object-aware visual-language space with massive data~\cite{detic, GLIP, grounding_dino, yao2022detclip} from various resources, including image classification datasets~\cite{deng2009imagenet}, object detection datasets~\cite{coco, gupta2019lvis, shao2019objects365, wang2023v3det}, grounding datasets~\cite{hudson2019gqa, plummer2015flickr30k} and image-text datasets~\cite{cc3m}, has shown impressive results. Further, multi-task learning with other language tasks, such as masked language modeling~\cite{zhang2022glipv2} and dense captioning~\cite{long2023capdet, yao2024detclipv3} can achieve better vision-language alignment, thus improving the detector's open-vocabulary ability. However, prior works~\cite{long2023capdet, yao2024detclipv3, lin2024generative, wu2025grit} only focus on generating short phrases for regions of interest. In this work, we explore another co-training task, \ie generating image-level detailed captions using large language models.

\begin{figure*}[t]
  \centering
  \vspace{-0.6em}
  \includegraphics[width=1\linewidth]{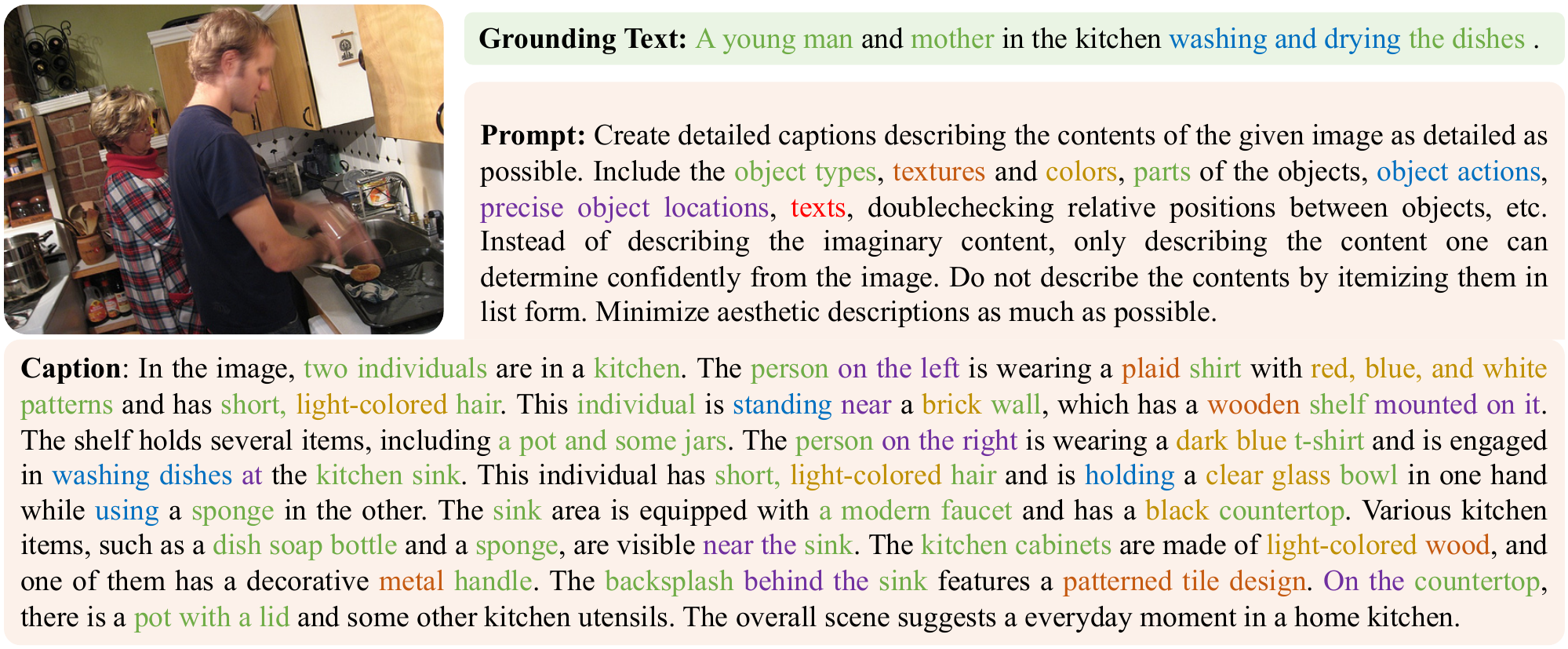}
  \caption{An example of GroundingCap-1M. Bounding box annotations are discarded for clarity. Compared with original short grounding texts, the detailed captions in GroundingCap-1M are rich in object types, textures, colors, parts of the objects, object actions, precise object locations and texts. Each caption in GroundingCap-1M has around 115 words on average.}
  \vspace{-0.6em}
  \label{fig:dataset_example}
\end{figure*}

\subsection{Large Vision-Language Model}

Recent large vision-language models~\cite{llava, liu2024improved, llava-onevision} equip large language models~\cite{touvron2023llama, vicuna2023, yang2024qwen2} with superior visual perception and understanding ability. A common large vision-language model contains three parts: a vision foundation models~\cite{clip, zhai2023sigmoid} to extract vision tokens, a projector to map vision features to the language space, and a large language model to understand both visual and text input. Recent works~\cite{tong2024cambrian} find that a better vision encoder improves the multi-modal performance of the final large vision-language model. \textbf{But whether the large language model can in turn improve the vision encoder is less explored.} InternVL~\cite{chen2024internvl} scales up a CLIP-like vision encoder to 6B parameters with a large language model as the text encoder. In this work, we show that the detector can also benefit from large language models and the improved detector can boost the multi-modal performance of the large language model, achieving mutual benefits.

To train a better large vision-language model, high-quality caption data is indispensable~\cite{llava-onevision, chen2023sharegpt4v, li2024monkey}. We argue that the quality of captions is also a key factor in training an open-vocabulary detector under the supervision of large language models. Thus, we take advantage of existing high-quality caption datasets and lead large vision-language models to generate high-quality data.

\section{GroundingCap-1M Dataset}

\noindent{\textbf{Data Formulation.}}
To support LLMDet training under the supervision of grounding loss and captioning loss, we formulate each training sample as a quadruple $(I, T_g, B, T_c)$, where $I$ is the image, $T_g$ is the short grounding text, $B$ are some annotated bounding boxes each of which is mapped to a phrase in the grounding text, and $T_c$ is the detailed caption for the whole image. An example is shown in \Cref{fig:dataset_example}. Two core principles are followed when collecting detailed captions for the whole image: \textbf{First, the caption should contain as many details as possible.} We expect the caption to describe object types, textures, colors, parts of the objects, object actions, precise object locations and texts in the image so that the caption is information-rich. \textbf{Second, the caption should only contain factual details about the image.} Too many imaginary or reasoning captions will reduce the information density or even hamper the model learning. The detailed and information-intensive captions will facilitate highly efficient training.

\begin{table}[t]
  \centering
  \resizebox{\linewidth}{!}{
      \begin{tabular}{lccc}
        \hline
        Dataset & Type & Caption Source & Size \\
        \hline
        COCO~\cite{coco} & Detection & ShareGPT4V~\cite{chen2023sharegpt4v}, ASv2~\cite{asv2} & 210k \\
        V3Det~\cite{wang2023v3det} & Detection & Our Caption (Qwen2-VL-72b~\cite{wang2024qwen2vl}) & 166k \\
        GoldG~\cite{GLIP} & Grounding & Our Caption (Qwen2-VL-72b~\cite{wang2024qwen2vl}) & 437k \\
        LCS~\cite{llava} & Image-Text & LLaVA-OneVision~\cite{llava-onevision}, ShareGPT4V~\cite{chen2023sharegpt4v} & 307k \\
        \hline
        \multicolumn{3}{l}{GroundingCap-1M} & 1120k \\
        \hline
      \end{tabular}
    }
  \vspace{-0.6em}
  \caption{Detailed dataset construction of GroundingCap-1M.}
  \vspace{-0.6em}
  \label{tab:dataset}
\end{table}

\vspace{0.5em}\noindent{\textbf{Dataset Construction.}} To save the construction cost, we start from existing datasets with either bounding boxes or detailed captions. Following previous works, the dataset is collected from object detection datasets, grounding datasets and image-text datasets.

For object detection datasets, we select well-known COCO~\cite{coco} and V3Det~\cite{wang2023v3det} datasets. As COCO is widely used in many multi-modal instruction tuning datasets, we can collect its detailed captions from existing assets. Specifically, we collect 168k captions from ShareGPT4V~\cite{chen2023sharegpt4v} which is known for detailed captions and 42k captions from ASv2~\cite{asv2} that mainly focus on object relations. V3Det is a dataset with 13k categories so that it can greatly enlarge the detector's vocabulary and is widely used in many open-vocabulary detectors~\cite{mm_GDINO, yao2024detclipv3}. The captions of V3Det are generated by us using Qwen2-VL-72b~\cite{wang2024qwen2vl} with the prompt presented in \Cref{fig:dataset_example}. Following GLIP~\cite{GLIP}, the grounding text of detection datasets is the concatenation of the class names in the dataset, \eg \textit{``chair. fork. cup. cow."}

\begin{figure}[t]
  \centering
  \vspace{-0.6em}
  \includegraphics[width=1\linewidth]{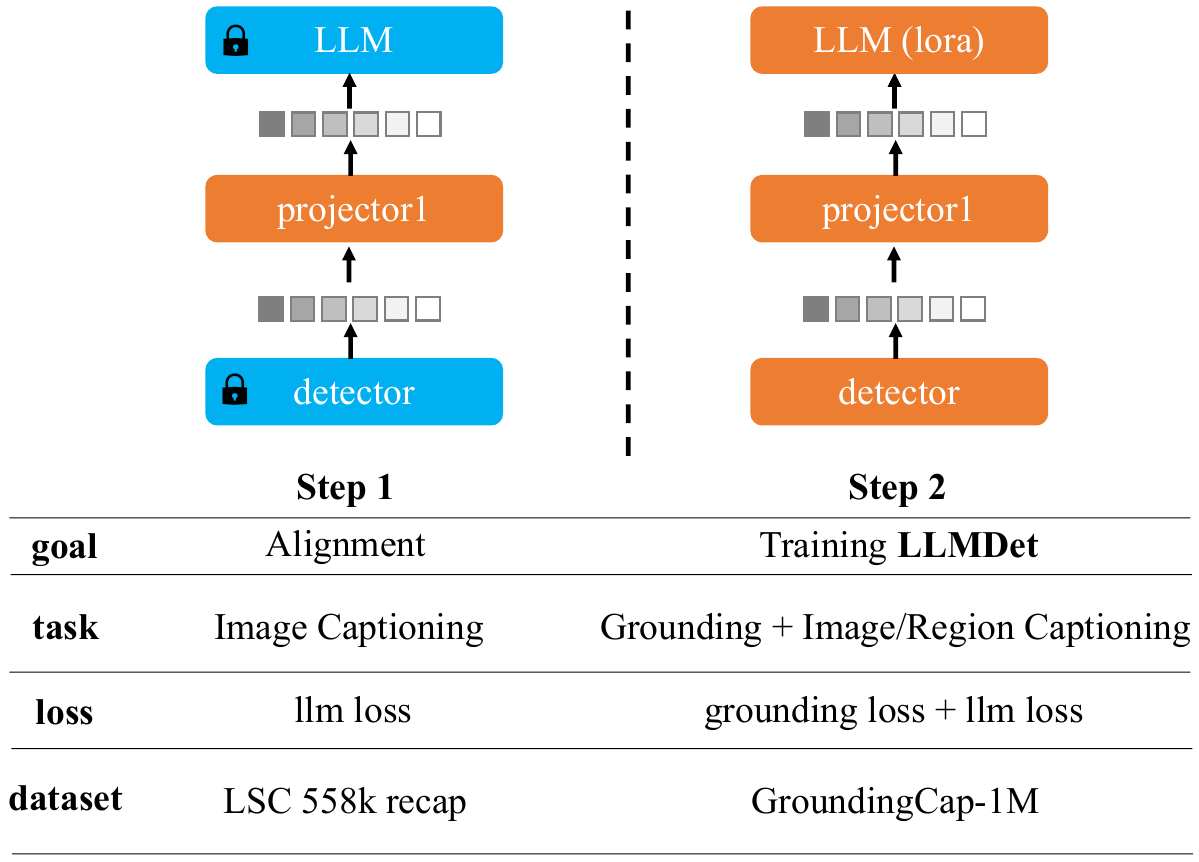}
  \caption{The multi-step training pipeline of LLMDet. In each step, modules in orange color are tunable while modules in blue color are frozen. In the first step, we train a projector to align the detector's features with the LLM so that we can integrate the LLM into the detector without breaking the pretrained features. Then, we train the detector with a standard grounding task and newly introduced captioning tasks in Step 2.}
  \vspace{-0.6em}
  \label{fig:pipeline}
\end{figure}

For grounding datasets, we choose widely-used GoldG~\cite{GLIP}, which contains GQA~\cite{hudson2019gqa} and Flickr30k~\cite{plummer2015flickr30k}. We find that the original annotations have many short grounding texts for each image. To save computation and increase negatives, we merge some grounding texts from the same image without bounding box conflicts into a single grounding text by simple concatenation. After merging, the dataset is down-sampled from 769k to 437k. The detailed caption is also generated by us using Qwen2-VL-72b~\cite{wang2024qwen2vl}.

For image-text datasets, LCS-558k~\cite{llava} with captions from LLaVA-OneVision~\cite{llava-onevision} and ShareGPT4V~\cite{chen2023sharegpt4v} is used. To generate pseudo boxes for images in this dataset, we first parse noun phrases from captions using a traditional language parser and then utilize MM\_Grounding\_DINO~\cite{mm_GDINO} (Swin-L) to generate bounding boxes for each phrase. The images with less than three bounding boxes are discarded. The grounding text is the concatenation of phrases in the same image, the same as detection datasets.

To sum up, the final dataset, \textbf{GroundingCap-1M}, contains 1120k samples, summarised in \Cref{tab:dataset}. 

\vspace{0.5em}\noindent\textbf{Quality Verification.} In the data collection procedure, we carefully select the prompts and use the best model (Qwen2VL-72b) we can access. A lot of work has been done to prevent hallucinations when training this top-performance model. However, it is inevitable that there will be some noise in the dataset. Thus we introduce some post-processing to clean the dataset. 1) We find that although we prompt the caption model not to describe the imaginary contents, the model still tends to output them but with some obvious words, like \textit{``indicating''}, \textit{``suggesting''}, \textit{``possibly''}. We simply delete the sub-sentences with speculative words. 2) We also design rules to filter out meaningless captions, \eg ``In the image, a man a man a man...(repeating)'' or ``Sorry, I can not answer the question.'' 3) To ensure the captions are rich with details, we use Qwen2VL-72b to re-generate captions for images whose first-time generated caption is less than 100 tokens. The double-check mechanism ensures the quality of the dataset.
After post-processing, each caption has around 115 words on average. \Cref{fig:dataset_example} shows an example from the GroundingCap-1M dataset. More examples can be found in Appendix. Some quantitive analyses are shown in \Cref{sec:ab}.

\section{Training LLMDet under the Supervision of Large Language Models}
\label{sec:training}

\begin{figure*}[t]
  \centering
  \includegraphics[width=0.9\linewidth]{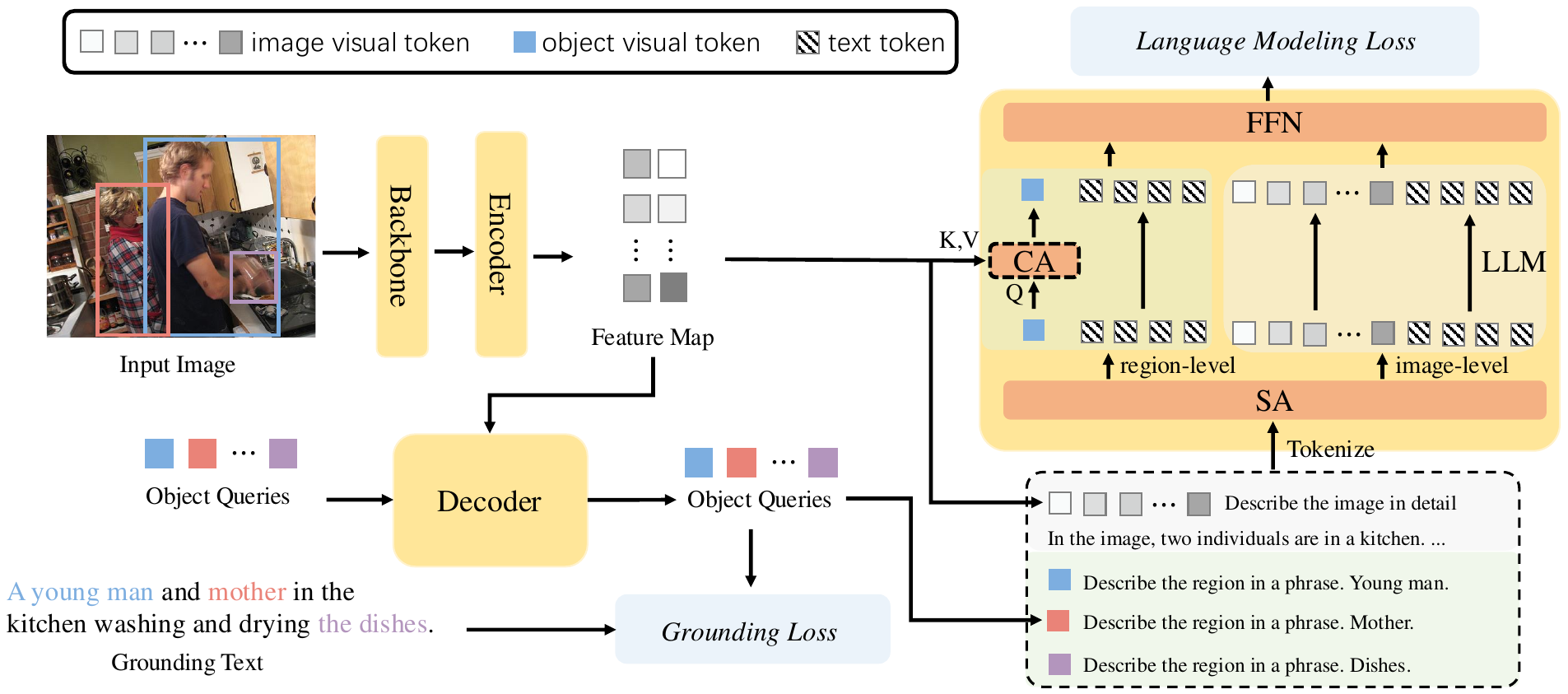}
  \caption{The overview of LLMDet. LLMDet contains a standard open-vocabulary detector and a large language model (LLM) and is trained under both grounding loss and language modeling loss. The LLM is designed to generate both \colorbox{image-level}{image-level captions} using feature maps as visual input and \colorbox{region-level}{region-level captions} using a single object query as visual input, which are separated by different prompts. Only vision tokens in region-level generation pass through the cross-attention (CA) modules in LLM, which is highlighted by a dashed boundary. Since the number of tokens in image-level and region-level generation varies greatly, we forward the LLM twice separately to save memory and computation. The LLM can be discarded in the inference time so that there is no extra cost.}
  \label{fig:model}
  \vspace{-0.6em}
\end{figure*}

Unifying the grounding task and some other language tasks can enrich vision features with language knowledge thus broadening vision concepts and achieving better vision-language alignment. Prior works~\cite{long2023capdet, wu2025grit, lin2024generative, yao2024detclipv3} mainly focus on dense captioning, in which the language model is designed to generate short captions or class names to describe the single region of interest. However, the details of single objects, the relations between objects and more information about foreground and background are overlooked but this information can be depicted in a single detailed image-level caption. In this work, we show that the region-level open-vocabulary object detector can also benefit from long detailed image-level captions under the supervision of large language models. The overall pipeline is shown in \Cref{fig:pipeline}.

Specifically, we utilize a large language model (LLM) to generate captions based on a pretrained DETR-based open-vocabulary detector. Since the detector and the LLM are pretrained separately, we first train a projector to map the vision features from the detector to the LLM's input space following common practices in training large multi-modal models. We take the p5 feature map from the detector's encoder as the LLM's input and the LLM is asked to generate the full image captions under the supervision of language modeling loss. Only the projector is tunable during this step (\textbf{Step 1} in \Cref{fig:pipeline}).

After pre-alignment, the detector, the projector and the LLM are finetuned in an end-to-end manner (\textbf{Step 2} in \Cref{fig:pipeline}). Except for the original grounding task, including the word-region alignment loss $\mathcal{L}_{align}$ and the box regression loss $\mathcal{L}_{box}$, we also introduce two tasks: image-level caption generation and region-level caption generation. Details are illustrated in \Cref{fig:model}.

In \textbf{image-level caption generation task}, the language model takes the feature maps from the detector as visual inputs and outputs the corresponding long detailed captions annotated in GroundingCap-1M. Following the common practices in training a large multi-modal model, we organize the input data of LLM in the conversation format, which includes system messages, user inputs, and answers. The user inputs contain the vision features from the detector and the prompt, \eg \textit{Describe the image in detail.} And the answers are the captions from GroundingCap-1M. The LLM aims to output the answers based on the user inputs under the supervision of the standard language modeling loss $\mathcal{L}_{lm}^{image}$. Since the output answers include various details and the comprehensive understanding of the image, these visual cues should be modeled in the visual features so that the LLM can minimize the training loss and generate the captions correctly.

However, since the LLM takes the whole feature map as input in image-level caption generation, it is hard for LLM to map the entities in the image-level captions back to a specific region in the whole image. For example, in Figure 2, the ``dishes'' is only a small part of the image and there are many dish-like objects in the image. Thus, we further introduce the \textbf{region-level caption generation task} as compensation, which introduces a prior for LLM to map the region with the corresponding word. In this task, we select the positive object queries from the detector, which are queries matched to ground truth boxes in the label assignment, and use the LLM to generate their corresponding grounding phrases separately, such as \textit{``young man''}, \textit{``mother''} and \textit{``dishes''} in the \Cref{fig:model}. Similar to image-level generation, the inputs of LLM are also formatted in conversations but with a different prompt to separate different types of inputs, \ie \textit{Describe the region in a phrase.} As the visual feature in a single object query is limited, we add some cross-attention layers in the LLM for object queries to gather necessary information from the detector's feature maps. Note that the text tokens and visual tokens in image-level generation do not pass through these cross-attention layers and these layers are trained from scratch. By outputting the corresponding phrases for object queries, the LLM can match the entities to a specific region exactly.

The overall training objective of LLMDet is the combination of the grounding loss and the generation losses:
\begin{equation}
    \mathcal{L} = \mathcal{L}_{align} + \mathcal{L}_{box} + \mathcal{L}_{lm}^{image} + \mathcal{L}_{lm}^{region}
\end{equation}
where $\mathcal{L}_{lm}^{region}$ is region-level caption generation loss.

\begin{table*}[t]
  \centering
  \vspace{-0.6em}
  \resizebox{\linewidth}{!}{
      \begin{tabular}{lll|cccc|cccc}
        \hline
        \multirow{2}{*}{Method} & \multirow{2}{*}{Backbone} & \multirow{2}{*}{Pre-training data} & \multicolumn{4}{c|}{LVIS$^{\text{minival}}$}  & \multicolumn{4}{c}{LVIS} \\
        & & & AP & AP$_{r}$ & AP$_{c}$ & AP$_{f}$ & AP & AP$_{r}$ & AP$_{c}$ & AP$_{f}$ \\
        \hline
        GLIP~\cite{GLIP} & Swin-T & O365,GoldG,Cap4M & 26.0 & 20.8 & 21.4 & 31.0 & 17.2 & 10.1 & 12.5 & 25.2 \\
        GLIPv2~\cite{zhang2022glipv2} & Swin-T & O365,GoldG,Cap4M & 29.0 & – & – & – & – & – & – & – \\
        CapDet~\cite{long2023capdet} & Swin-T & O365,VG & 33.8 & 29.6 & 32.8 & 35.5 & – & – & – & – \\
        Grounding-DINO~\cite{grounding_dino} & Swin-T & O365,GoldG,Cap4M & 27.4 & 18.1 & 23.3 & 32.7 & 20.1 & 10.1 & 15.3 & 29.9 \\
        OWL-ST~\cite{OWL-ST} & CLIP B/16 & WebLI2B & 34.4 & 38.3 & – & – & 28.6 & 30.3 & – & – \\
        Desco-GLIP~\cite{li2024desco} & Swin-T & O365,GoldG,CC3M & 34.6 & 30.8 & 30.5 & 39.0 & 26.2 & 19.6 & 22.0 & 33.6 \\
        DetCLIP~\cite{yao2022detclip} & Swin-T & O365,GoldG,YFCC1M & 35.9 & 33.2 & 35.7 & 36.4 & 28.4 & 25.0 & 27.0 & 28.4 \\
        DetCLIPv2~\cite{yao2023detclipv2} & Swin-T & O365,GoldG,CC15M & 40.4 & 36.0 & 41.7 & 40.4 & 32.8 & 31.0 & 31.7 & 34.8 \\
        DetCLIPv3~\cite{yao2024detclipv3} & Swin-T & O365,V3Det,GoldG,GranuCap50M & 47.0 & 45.1 & 47.7 & 46.7 & 38.9 & 37.2 & 37.5 & 41.2 \\
        YOLO-World-L~\cite{yolo_world} & YOLOv8-L & O365,GoldG,CC3M & 35.4 & 27.6 & 34.1 & 38.0  & –  & –  & –  & – \\
        T-Rex2~\cite{T-Rex2} & Swin-T & 10M data from various resources & 42.8 & 37.4 & 39.7 & 46.5 & 34.8 & 29.0 & 31.5 & 41.2 \\
        OV-DINO~\cite{ov-dino} & Swin-T & O365,GoldG,CC1M & 40.1 & 34.5 & 39.5 & 41.5 & 32.9 & 29.1 & 30.4 & 37.4 \\
        \hline
        MM-GDINO~\cite{mm_GDINO} & Swin-T & O365,GoldG,GRIT,V3Det & 41.4 & 34.2 & 37.4 & 46.2 & 31.9 & 23.6 & 27.6 & 40.5 \\
        \rowcolor{ours} LLMDet & Swin-T & GroundingCap-1M & 44.7 & 37.3 & 39.5 & 50.7 & 34.9 & 26.0 & 30.1 & 44.3 \\
        \hline
        \hline
        GLIP~\cite{GLIP} & Swin-L & FourODs,GoldG,Cap24M & 37.3 & 28.2 & 34.3 & 41.5 & 26.9 & 17.1 & 23.3 & 36.4 \\
        GLIPv2~\cite{zhang2022glipv2} & Swin-H & FiveODs,GoldG,CC15M,SBU & 50.1 & – & – & – & – & – & – & – \\
        Grounding-DINO~\cite{grounding_dino} & Swin-L & O365,OI,GoldG,Cap4M,COCO,RefC & 33.9 & 22.2 & 30.7 & 38.8 & – & – & – & – \\
        OWL-ST~\cite{OWL-ST} & CLIP L/14 & WebLI2B & 40.9 & 41.5 & – & – & 35.2 & 36.2 & – & – \\
        DetCLIP~\cite{yao2022detclip} & Swin-L & O365,GoldG,YFCC1M & 38.6 & 36.0 & 38.3 & 39.3 & 28.4 & 25.0 & 27.0 & 31.6 \\
        DetCLIPv2~\cite{yao2023detclipv2} & Swin-L & O365,GoldG,CC15M & 44.7 & 43.1 & 46.3 & 43.7 & 36.6 & 33.3 & 36.2 & 38.5 \\
        DetCLIPv3~\cite{yao2024detclipv3} & Swin-L & O365,V3Det,GoldG,GranuCap50M & 48.8 & \textbf{49.9} & \textbf{49.7} & 47.8 & 41.4 & \textbf{41.4} & \textbf{40.5} & 42.3 \\
        \hline
        MM-GDINO~\cite{mm_GDINO} & Swin-B & O365,GoldG,V3Det & 44.5 & 37.5 & 39.9 & 49.9 & 34.9 & 26.7 & 30.4 & 43.5 \\
        MM-GDINO~\cite{mm_GDINO} & Swin-L & O365V2,OpenImageV6,GoldG & 36.8 & 28.1 & 31.8 & 42.8 & 29.1 & 19.7 & 25.6 & 37.2 \\
        \rowcolor{ours} LLMDet & Swin-B & GroundingCap-1M & 48.3 & 40.8 & 43.1 & 54.3 & 38.5 & 28.2 & 34.3 & 47.8 \\
        \rowcolor{ours} LLMDet & Swin-L & GroundingCap-1M & \textbf{51.1} & 45.1 & 46.1 & \textbf{56.6} & \textbf{42.0} & 31.6 & 38.8 & \textbf{50.2} \\
        \hline
      \end{tabular}
    }
  \vspace{-0.6em}
  \caption{Zero-shot fixed AP~\cite{dave2021evaluating} on LVIS val~\cite{gupta2019lvis} and minival~\cite{kamath2021mdetr}. LLMDet achieves state-of-the-art performance with much less data.}
  \vspace{-0.6em}
  \label{tab:lvis_result}
\end{table*}

\begin{table}[t]
  \centering
  \resizebox{\linewidth}{!}{
      \begin{tabular}{lccc}
        \hline
        Model & Backbone & ODinW13 & ODinW35 \\
        \hline
        MDETR~\cite{kamath2021mdetr} & ENB5 & - & 10.7 \\
        T-Rex2~\cite{T-Rex2} & Swin-T & - & 18.0 \\
        OWL-ViT~\cite{OWL-ST} & ViT L/14(CLIP) & \textbf{53.0} & 18.8 \\
        GLIP~\cite{GLIP} & Swin-T & 46.5 & 19.6 \\
        GLIPv2~\cite{zhang2022glipv2} & Swin-T & 48.5 & 22.3 \\
        DetCLIP~\cite{yao2022detclip} & Swin-T & 43.3 & - \\
        Grounding-DINO~\cite{grounding_dino} & Swin-T & 51.4 & 22.7 \\
        \hline
        MM-GDINO~\cite{mm_GDINO} & Swin-T & 52.5 & 23.1 \\
        \rowcolor{ours} LLMDet  & Swin-T & 52.1 & \textbf{23.8} \\
        \hline
      \end{tabular}
    }
  \vspace{-0.6em}
  \caption{Zero-shot transfer on ODinW~\cite{li2022elevater}.}
  \label{tab:odinw}
\end{table}

\section{Experiment}

\subsection{Implementation Details}
In this work, we select MM\_Grounding\_DINO~\cite{mm_GDINO} (MM-GDINO for short in the following) as the baseline model since it is fully open-sourced and enjoys SOTA performance. We simply reload their pretrained checkpoint and finetune the model with our GroundingCap-1M dataset under the supervision of both the grounding loss and the caption generation loss. Note that a large part of images in GroundingCap-1M is the same as the pretraining datasets used in MM-GDINO, such as GoldG~\cite{GLIP} and V3Det~\cite{wang2023v3det}. Since MM-GDINO is fully-pretrained, the vision backbone of MM-GDINO is frozen during training. The large language model is initialized from LLaVA-OneVision-0.5b-ov~\cite{llava-onevision}. To save memory and improve training efficiency, we set the maximum token length for the image-level generation as 1600 and the one for region-level generation as 40. The maximum number of regions for caption generation per image is limited to 16. For image-level visual input, we use the p4 and p5 feature maps from the detector's encoder. We resize p4 to $27\times27$ and p5 to $20\times20$ and concatenate them as a single token sequence. We implement LLMDet in MMDetection~\cite{mmdetection} using automatic mixed-precision and gradient checkpointing and train it for 150k iterations (around two epochs) with batch size 16, which can be done around two days on eight NVIDIA L20 GPUs.

\begin{table}[t]
  \centering
  \resizebox{\linewidth}{!}{
      \begin{tabular}{lcccc}
        \hline
        \multirow{2}{*}{Method} & \multirow{2}{*}{Backbone} & COCO~\cite{coco} & COCO-O~\cite{mao2023coco} & Effective \\
        & & AP & AP & Robustness \\
        \hline
        DINO~\cite{zhang2022dino} & R50 & \textcolor{gray}{49.0} & 22.5 & +0.5 \\
        GLIP~\cite{GLIP} & Swin-T & 46.1 & 29.0 & +8.0 \\
        DetCLIPv3~\cite{yao2024detclipv3} & Swin-T & 47.2 & \textbf{38.5} & \textbf{+17.3} \\
        Grounding-DINO~\cite{grounding_dino} & Swin-T & 48.4 & 37.6 & +15.8 \\
        \hline
        MM-GDINO~\cite{mm_GDINO} & Swin-T & 50.4 & 34.0 & +11.3 \\
        \rowcolor{ours} LLMDet  & Swin-T & \textcolor{gray}{55.6} & 36.1 & +11.1 \\
        \hline
      \end{tabular}
    }
  \vspace{-0.6em}
  \caption{Distribution shift performance on COCO-O dataset. \textcolor{gray}{Gray} numbers indicate including COCO data in training.}
  \label{tab:coco-o}
\end{table}

\subsection{Zero-Shot Detection Transfer Ability}

\begin{table*}[t]
  \centering
  \resizebox{\linewidth}{!}{
      \begin{tabular}{l|c|ccc|ccc|cc|ccc}
        \hline
        \multirow{2}{*}{Method} & \multirow{2}{*}{Backbone}  & \multicolumn{3}{c|}{RefCOCO~\cite{refcoco}} & \multicolumn{3}{c|}{RefCOCO+~\cite{refcoco+}} & \multicolumn{2}{c|}{RefCOCOg~\cite{refcocog}} & \multicolumn{3}{c}{gRefCOCO~\cite{grefcoco}} \\
        & & val & testA & testB & val & testA & testB & val & test & val & testA & testB \\
        \hline
        GLIP (B)~\cite{GLIP} & Swin-T & 50.0 & 54.7 & 43.1 & 49.0 & 53.4 & 43.4 & 65.6 & 66.1 & - & - & - \\
        GLIP~\cite{GLIP} & Swin-T & 50.4 & 54.3 & 43.8 & 49.5 & 52.8 & 44.6 & 66.1 & 66.9 & - & - & - \\
        Grounding-DINO~\cite{grounding_dino} & Swin-T & 50.8 & 57.4 & 45.0 & 51.6 & 57.3 & 46.4 & 60.4 & 59.7 & 40.5/83.8 & \textbf{29.3}/82.9 & 30.0/86.1 \\
        \hline
        MM-GDINO~\cite{mm_GDINO} & Swin-T & 66.0 & 70.3 & 60.0 & 54.3 & 60.4 & 49.2 & \textbf{72.6} & \textbf{72.5} & \textbf{41.0}/\textbf{91.3} & 26.1/\textbf{93.0} & 30.4/92.3 \\
        \rowcolor{ours} LLMDet  & Swin-T & \textbf{69.5} & \textbf{73.0} & \textbf{64.0} & \textbf{55.3} & \textbf{60.6} & \textbf{49.3} & 72.1 & \textbf{72.5} & 40.7/91.2 & 26.2/92.7 & \textbf{30.5}/\textbf{92.9} \\
        \hline
      \end{tabular}
    }
  \vspace{-0.6em}
  \caption{Zero-shot transfer on common referring expression comprehension datasets. The evaluation metric for RefCOCO, RefCOCO+, and  RefCOCOg is the Top-1 accuracy. The evaluation metrics for gRefCOCO are Pr(F1=1, IoU$\geq$0.5) and N-acc.}
  \label{tab:ref}
\end{table*}

To demonstrate the great open-vocabulary ability of LLMDet, we select a wide range of benchmarks, including LVIS~\cite{gupta2019lvis}, ODinW13/35~\cite{li2022elevater}, COCO-O~\cite{mao2023coco}, RefCOCO~\cite{refcoco}, RefCOCO+~\cite{refcoco+}, RefCOCOg~\cite{refcocog}, gRefCOCO~\cite{grefcoco} and perform zero-shot testing on them. Since we use COCO~\cite{coco} dataset during training, we carefully remove the images in the RefCOCO/RefCOCO+/RefCOCOg validation and test set from GroundingCap-1M following MM-GDINO. The images in the LVIS minival are not overlapped with the training set of COCO so it strictly follows the zero-shot setting. During the test, the LLM is discarded, thus the inference cost is the same as our baseline.

\vspace{0.3em}\noindent{\textbf{Zero-shot performance on LVIS.}} LVIS~\cite{gupta2019lvis} is a detection dataset with 1203 classes. Based on the frequency, the classes can be divided into frequent, common and rare classes. Following GLIP, the total 1203 classes are split into 31 chunks, each with 40 classes. Thus the detector will be forwarded 31 times for each image. Note that a larger chunk size will improve the performance. As shown in \Cref{tab:lvis_result}, with the novel training objective and the new dataset, LLMDet outperforms the baseline MM-GDINO by 3.3\%/3.8\%/14.3\% AP and 3.1\%/3.3\%/17.0\% AP$_r$ on LVIS minival across different backbones. We find that the performance of MM-GDINO with Swin-L~\cite{liu2021swin} as the backbone is extremely low, which is probably due to different pretraining data, especially lacking V3Det. But LLMDet with Swin-L as the backbone still outperforms other SOTA methods with much less training data and achieves 50.6\% AP, showing great open-vocabulary capabilities. The same trend can be found on LVIS val. We notice that the DetCLIP series achieves more balanced performance on different classes, which is probably due to carefully collected and annotated datasets and a well-organized noun concept corpus. We believe LLMDet can also be applied to DetCLIP.

\vspace{0.3em}\noindent{\textbf{Zero-shot performance on ODinW.}} ODinW (Object Detection in the Wild)~\cite{li2022elevater} is a collection of 35 datasets across various domains and vocabularies, which is a challenging and comprehensive benchmark for open-vocabulary detection. Following previous works, we report the average AP on selected 13 datasets (ODinW13) and all 35 datasets (ODinW35). LLMDet gets the highest AP on ODinW35, demonstrating the great transferring ability to a wide range of datasets. Detailed performance on each dataset can be found in Appendix.

\vspace{0.3em}\noindent{\textbf{Zero-shot performance on COCO-O.}} COCO-O~\cite{mao2023coco} is a dataset sharing the same 80 classes with COCO but with different domains, \ie sketch, weather, cartoon, painting, tattoo, and handmake, which are significantly different from natural images. Although the performance on COCO-O is highly related to the pretraining datasets, LLMDet still outperforms MM-GDINO by 2.1\% AP, showing that LLMDet is more robust to domain shifts. Detailed performance on each domain can be found in Appendix.

\vspace{0.3em}\noindent{\textbf{Zero-shot performance on referring expression comprehension datasets.}} Referring expression comprehension (REC) is a task to localize the objects referred by phrases, which needs comprehensive language understanding and fine-grained vision-language alignment. By co-training with LLMs using detailed captions, LLMDet can model rich visual details with enriched vision-language alignment. Thus LLMDet outperforms the baseline MM-GDINO on various REC datasets.


\subsection{Ablation Study}
\label{sec:ab}

In this subsection, experiments are conducted on the Swin-T backbone and report the performance on LVIS minival. Visualizations can be found in Appendix.

\begin{table}[t]
  \centering
  \resizebox{\linewidth}{!}{
      \begin{tabular}{ccc|cccc}
        \hline
        grounding & region-level & image-level & AP & AP$_{r}$ & AP$_{c}$ & AP$_{f}$ \\
        data & generation  & generation \\
        \hline
         &  &  & 41.4 & 34.2 & 37.4 & 46.2 \\
        \checkmark &  &  & 43.8 & 33.4 & 38.5 & 50.3 \\
        \checkmark & \checkmark &  & 43.7 & 34.2 & 38.1 & 50.3 \\
        \checkmark &  & \checkmark & 44.0 & 33.4 & 39.4 & 50.1 \\
        \rowcolor{ours} \checkmark & \checkmark & \checkmark & 44.7 & 37.3 & 39.5 & 50.7 \\
        \hline
      \end{tabular}
    }
  \vspace{-0.6em}
  \caption{Ablations on main components.}
  \vspace{-0.6em}
  \label{tab:ab1}
\end{table}

\vspace{0.3em}\noindent{\textbf{Effect of the main components of LLMDet.}} In this work, we collect a new dataset GroundingCap-1M, which contains both grounding annotations and detailed long captions for each image. As shown in \Cref{tab:ab1}, finetuning with only grounding annotations can boost the performance from 41.4\% AP to 43.8\% AP. We also show that with only region-level generation, the performance can not be improved since the region-level captions in LLMDet are just the class names or grounding phrases of regions, which do not provide extra information. Simply using image-level generation can slightly improve the performance. As explained in \Cref{sec:training}, the LLM may find it hard to map the entities back to a specific object from a whole image. Thus, combining both image-level and region-level generation can fully unleash the benefit of the LLM's supervision signals. And the rich vision-language representations learned from detailed captions significantly improve 3.9\% AP$_r$ (Row 2 vs Row 5), showing that the fine-grained visual representations are helpful in recognizing rare classes.

\begin{table}[t]
  \centering
  \resizebox{\linewidth}{!}{
      \begin{tabular}{l|cccc}
        \hline
        LLM & AP & AP$_{r}$ & AP$_{c}$ & AP$_{f}$ \\
        \hline
        Qwen2-0.5b-instruct~\cite{yang2024qwen2} & 44.4 & 36.4 & 39.2 & 50.5  \\
        \rowcolor{ours} LLaVA-OneVision-0.5b-ov~\cite{llava-onevision} & 44.5 & 38.6 & 39.3 & 50.3 \\
        Qwen2-1.5b-instruct~\cite{yang2024qwen2} & 44.6 & 35.3 & 39.5 & 50.8 \\
        \hline
      \end{tabular}
    }
  \vspace{-0.6em}
  \caption{Ablations on large language models.}
  \vspace{-0.6em}
  \label{tab:ab3}
\end{table}

\vspace{0.3em}\noindent{\textbf{Effect of different large language models.}} By default, we use the LLM in LLaVA-OneVision-0.5b-ov~\cite{llava-onevision}, which is finetuned from Qwen2-0.5b-instruct~\cite{yang2024qwen2}. Since the LLM in LLaVA-OneVision-0.5b-ov is pretrained with abundant multi-modal data but with a different vision encoder, the pretraining can still improve the performance, especially for rare classes (+2.2\% AP$_r$), as shown in \Cref{tab:ab3}. But we find that increasing the size of the LLM only slightly improves the performance, perhaps larger language models mainly improve in reasoning ability which does not benefit the detector's visual representations.

\begin{table}[t]
  \centering
  \resizebox{\linewidth}{!}{
      \begin{tabular}{c|l|c|c|c|cccc}
        \hline
        Exp & Captions & len & Det.$\uparrow$ & Hul.$\downarrow$ & AP & AP$_{r}$ & AP$_{c}$ & AP$_{f}$ \\
        \hline
        \rowcolor{ours} 1 & GroundingCap-1M & 115 & 4.63 & 1.15 & 44.5 & 38.6 & 39.3 & 50.3 \\
        2 & LLaVA-Onevision-7B generated & 105 & 4.37 & 1.34 & 43.6 & 33.5 & 38.0 & 50.5 \\
        3 & Original captions / grounding texts & 30 & 3.04 & 0.90 & 43.2 & 35.7 & 37.3 & 49.9 \\
        \hline
      \end{tabular}
    }
  \vspace{-0.6em}
  \caption{Ablations on the quality of generated captions. Detailedness (Det) and hallucination (Hul) scores are measured by GPT-4o ranging from 0 to 5. Len is the average length of captions.}
  \vspace{-0.6em}
  \label{tab:r1}
\end{table}

\vspace{0.3em}\noindent{\textbf{Effect of generated captions' quality.}} As shown in \Cref{tab:r1}. We first replace the captions generated by Qwen2VL-72b with the ones from LLaVA-Onevision-7B, including captions in V3Det, GoldG and part of LCS. The performance significantly decreases by 0.9\% AP and 5.1\% AP$_{r}$. We further replace our generated captions with COCO captions, LCS captions from LLaVA, and short grounding texts in GoldG. The performance further decreases by 0.4\% AP. To directly compare the detailedness and the degree of hallucinations in generated captions, we randomly sample 100 captions from each part of the datasets (detection, grounding, and image-text pair), a total of 300 captions per experiment. And we utilize GPT-4o~\cite{hurst2024gpt} as a judge to give a comprehensive score for each caption-image pair. The used prompts are shown in Appendix. The captions in GroundingCap-1M have the highest detailedness scores and moderate hallucinations, validating the superior quality of our dataset. As human-annotated captions have fewer hallucinations (0.90 vs 1.34, LLaVA captions still have hallucinations), the AP$_{r}$ in Exp 3 is even higher than the one in Exp 2.

\vspace{0.3em}\noindent{\textbf{Effect of the pretraining data.}} In this work, we collect the GroundingCap-1M dataset. Due to computation constraints, the dataset only contains 1M data, which is much less than the datasets used in other open-vocabulary detectors. As shown in the second row in \Cref{tab:ab4}, if we do not use the LCS dataset in GroundingCap-1M (813k data now), the performance significantly reduces to 42.8\% AP, showing that more pretraining data will further improve LLMDet's performance. Further, the image-level captions should only contain factual details about the image so that we utilize a post-processing procedure to delete the sub-sentences with speculative words. If we do not delete them, the performance drops to 44.2\% AP and 35.0\% AP$_r$, showing that hallucinations may significantly affect rare class performance.

\vspace{0.3em}\noindent{\textbf{Effect of the cross-attention layers in LLM.}} In LLMDet, visual tokens in region-level generation pass through cross-attention layers, while text tokens and visual tokens in image-level generation do not pass through them. We ablate the design in the third and fourth rows of \Cref{tab:ab4}. Visual tokens in the region-level generation are single object queries that contain little visual information. If object queries do not gain necessary information from the detector's encoder feature maps through cross-attention, the performance will degrade to 44.0\% AP. We further find that using cross-attention in image-level generation is not helpful as we use the whole feature maps in image-level generation.

\vspace{0.3em}\noindent{\textbf{Effect of pretraining the projector before end-to-end finetuning.}} LLMDet improves the rare class performance by pursuing fine-grained vision-language alignment through co-training with LLMs using high-quality captions. Since the LLM and the detector are pretrained separately, pretraining the projector makes their feature space aligned while preserving the pretrained knowledge. Without pretraining the projector, it affects the alignment and decreases the rare class AP (-3.5\% AP$_r$), as shown in the last row of \Cref{tab:ab4}. As frequent classes have abundant annotations, the negative impacts can be alleviated.


\begin{table}[t]
  \centering
  \resizebox{\linewidth}{!}{
      \begin{tabular}{c|l|cccc}
        \hline
        Type & Method & AP & AP$_{r}$ & AP$_{c}$ & AP$_{f}$ \\
        \hline
        \rowcolor{ours} - & LLMDet & 44.5 & 38.6 & 39.3 & 50.3  \\
        \hline
        \multirow{2}{*}{data} & Do not use LCS dataset & 42.8 & 33.7 & 37.7 & 48.9 \\
        & Do not delete imaginary sentences & 44.2 & 35.0 & 38.9 & 50.6 \\
        \hline
        \multirow{2}{*}{archi} & Do not use CA in R-L generation & 44.0 & 35.1 & 38.7 & 50.2 \\
         & Use CA in both R-L and I-L generation & 44.4 & 36.6 & 39.4 & 50.2 \\
        \hline
        pipeline & Do not pretrain the projector & 44.4 & 35.1 & 39.8 & 50.3 \\
        \hline
      \end{tabular}
    }
  \vspace{-0.6em}
  \caption{More ablations on the data, architecture, and pipeline. R-L and I-L denotes region level and image level.}
  \vspace{-0.6em}
  \label{tab:ab4}
\end{table}

\section{Conclusion}

In this work, we explore a new training objective to boost the performance of existing open-vocabulary detectors. By utilizing a large language model to generate both image-level detailed captions and region-level coarse grounding phrases, the detector receives more information and a more comprehensive understanding of the image from the detailed captions and builds rich vision-language representations. The resulting detector, LLMDet, achieves state-of-the-art performance across a wide range of benchmarks. We also show that the improved LLMDet can in turn build a strong large multi-modal model, achieving mutual benefits. We hope our work can provide insights into enhancing vision models with top-performed large language models.

{
    \small
    \bibliographystyle{ieeenat_fullname}
    \bibliography{main}
}

\newpage
\mbox{}
\newpage

\begin{appendices}
\setcounter{table}{0} 
\setcounter{figure}{0} 
\setcounter{equation}{0} 
\renewcommand{\thetable}{\thesection-\arabic{table}} 
\renewcommand{\theequation}{\thesection-\arabic{equation}} 
\renewcommand{\thefigure}{\thesection-\arabic{figure}} 

\section{LLMDet Builds a Stronger Large Vision-Language Model}

\begin{figure}[t]
  \centering
  \vspace{-0.6em}
  \includegraphics[width=1\linewidth]{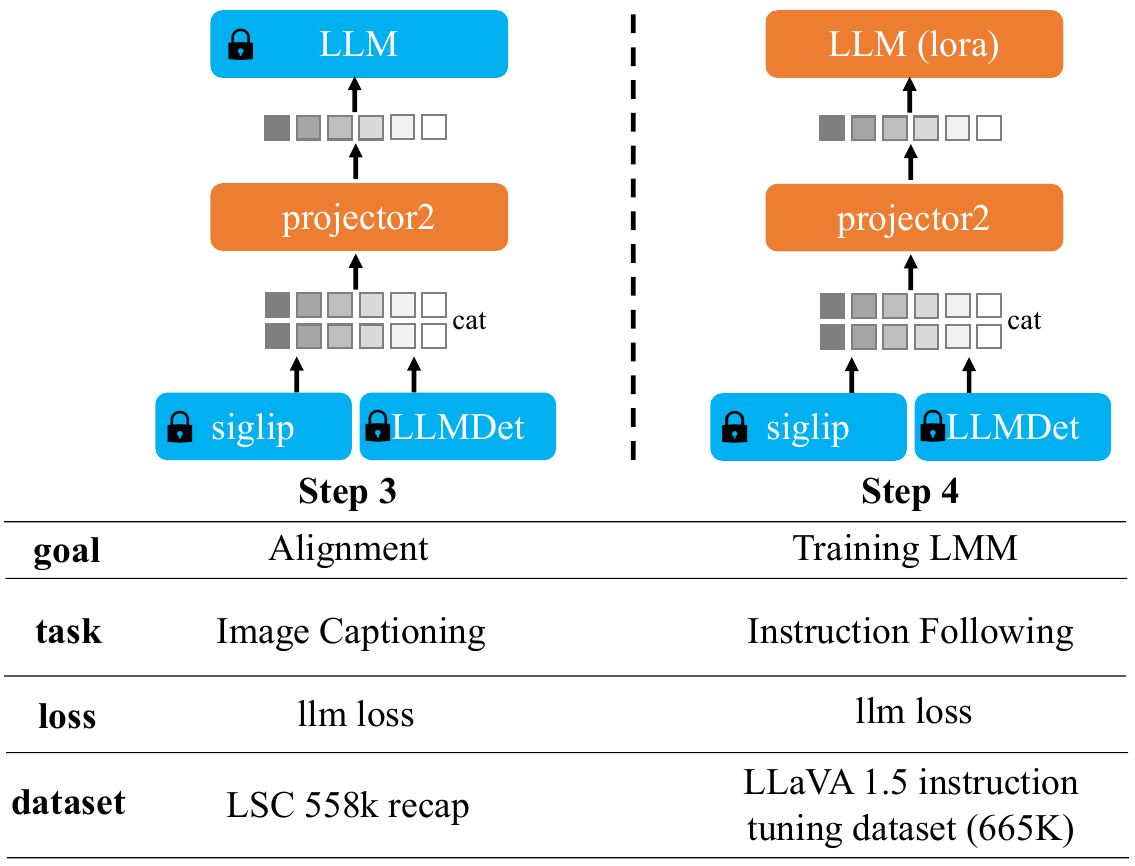}
  \caption{The multi-step training pipeline of using LLMDet to build a strong large multi-modal model. The large multi-modal model uses a mixture of vision encoders, including LLMDet and SigLIP. In each step, modules in orange color are tunable while modules in blue color are frozen. We first pretrain a new projector and then finetune the large multi-modal model with visual instruct tuning.}
  \vspace{-0.6em}
  \label{fig:pipeline2}
\end{figure}

In this subsection, we show that LLMDet can serve as a general vision foundation model and in turn gets a strong large multi-modal model. Recent large multi-modal models (LMM) are based on pretrained large language models and pretrained vision foundation models. Different vision foundation models will significantly affect the performance of LMMs~\cite{zong2024mova}. Since LLMDet is enhanced under the supervision of long detailed image-level captions and pre-aligned with LLM, LLMDet inherits great potential to build a stronger LMM. Following recent advances~\cite{zong2024mova, shi2024eagle, tong2024cambrian}, we build the LMM using a mixture of vision experts, \ie a SigLIP~\cite{zhai2023sigmoid} vision encoder and our LLMDet. As shown in \Cref{fig:pipeline2}, the visual features from two vision encoders are concatenated along the channel dimension, and then a projector is utilized to map the features to the LLM's input space. We start from LLaVA-OneVision-0.5b-ov~\cite{llava-onevision} and insert our LLMDet to it as shown in \Cref{fig:pipeline2}. We first pretrain a new projector and then finetune the LLM with the LLaVA 1.5~\cite{liu2024improved} instruction tuning dataset which is only a small part of the dataset used in LLaVA-Onevision. 

We select three representative benchmarks to evaluate the multi-modal performance of the LMM: the comprehensive understanding benchmark MME~\cite{fu2023mme}, the hallucination benchmark POPE~\cite{pope} and the academic VQA benchmark GQA~\cite{hudson2019gqa}. As shown in \Cref{tab:llm}, combining the MM-GDINO to LLaVA-OneVision-0.5b-ov can improve the performance on GQA and POPE. As detectors excel at localizing objects in the image, the precise localization makes the LLM aware of the objects existed in images, which helps the LLM overcome hallucination and perform simple QA about objects in the image. The multi-modal perception and understanding ability can be further enhanced with a stronger LLMDet which is also pre-aligned~\cite{shi2024eagle} with the LLM in LLaVA-OneVision-0.5b-ov. The resulting LMM achieves the highest performance on the MME benchmark, validating the mutual benefits between the detector and the LMM.

\begin{table}[t]
  \centering
  \resizebox{\linewidth}{!}{
      \begin{tabular}{l|c|ccc|cc}
        \hline
        \multirow{2}{*}{Method} & GQA & \multicolumn{3}{c|}{POPE~\cite{pope}} & \multicolumn{2}{c}{MME~\cite{fu2023mme}} \\
        & \cite{hudson2019gqa} & rand & pop & adv & perception & cognition \\
        \hline
        OneVision-0.5b & 56.9 & 87.5 & 86.3 & 85.0 & 1238 & 240 \\
        \hline
        OneVision-0.5b & \multirow{2}{*}{\textbf{61.2}} & \multirow{2}{*}{\textbf{88.9}} & \multirow{2}{*}{\textbf{88.1}} & \multirow{2}{*}{\textbf{86.6}} & \multirow{2}{*}{1207} & \multirow{2}{*}{256}  \\
        +MM-GDINO & & & & \\
        \hline
        \rowcolor{ours} OneVision-0.5b &  &  &  &  &  &   \\
        \rowcolor{ours} +LLMDet & \multirow{-2}{*}{\textbf{61.2}} & \multirow{-2}{*}{88.8} & \multirow{-2}{*}{88.0} & \multirow{-2}{*}{86.0} & \multirow{-2}{*}{\textbf{1297}} & \multirow{-2}{*}{\textbf{264}} \\
        \hline
      \end{tabular}
    }
  \caption{Multi-modal performance using different vision encoders. OneVision-0.5b is short for LLaVA-OneVision-0.5b-ov~\cite{llava-onevision}.}
  \label{tab:llm}
\end{table}

\section{Limitations}

Although we provide detailed captions to train LLMs, we find that the LLM co-trained with detectors tends to output relatively short descriptions for the whole image, even given the prompts to describe the image in detail. We suppose the reason is that our region-level data is far more than the image-level data (one image has multiple regions). 

Further, our region-level descriptions are too simple as they are just the grounding phrases of the regions. We believe collecting some high-informative data for regions like DetCLIPv3 can further improve the performance.

\section{Implement Details of Zero-Shot Test on Referring Expression Comprehension Datasets}

In this work, LLMDet is trained with phrase grounding loss and caption generation loss. In the phrase grounding task, the model is asked to detect each phrase in the given grounding text. For example, the model is expected to detect ``the man'' and ``umbrella'' in the text ``the man with an umbrella''.

To demonstrate the great open-vocabulary ability of LLMDet, we directly transfer LLMDet to the referring expression comprehension (REC) task, which is a task slightly different from the phrase grounding task. In REC, the model should only detect the single object referred by the given sentence. For example, the model should only detect ``the man'' in the text ``the man with an umbrella'', which means discrepancies exist between the pretraining task and the target task. Thus, we find that the model tends to predict the ``umbrella'' with the highest confidence. To minimize the discrepancies, we first use NLTK~\cite{bird2006nltk} tools to find the subject in the text and then select the box with the highest confidence corresponding to the subject as the answer.

\section{Prompts for Calculating Detailedness and Hallucination Scores}

In \Cref{tab:r1}, we utilize GPT-4o as a judge to give a comprehensive score for each caption-image pair. We referred to HalluciDoctor~\cite{yu2024hallucidoctor} and adopted similar prompts as follows.

\begin{tcolorbox}[colback=gray!10,colframe=black!80,title=The prompt for calculating hallucination scores]
Suppose you are a hallucination annotator who judges the degree of hallucination based on the number of errors in the description of objects, relations, and attributes. You should check each sentence in the description one by one.
\\

\{image\}

Please carefully compare the image and the given caption below and provide the hallucination score (an integer value between 0 and 5) based on overall hallucinations in each sub-sentence, where the fewer descriptive errors in the caption, the lower the hallucination score given. Only output the score without any explanation. 

Description: \{caption\} 

Output:
\end{tcolorbox}

\begin{tcolorbox}[colback=gray!10,colframe=black!80,title=The prompt for calculating detailedness scores]
Suppose you are an image detail annotator who judges the degree of sentence detailedness based on the object types, textures and colors, parts of the objects, object actions, precise object locations, and texts.
\\

\{image\}

Please carefully compare the image and the given caption below and provide the detailedness score (an integer value between 0 and 5) without any explanation, where caption with more factual content give a higher detailedness score. Only output the score without any explanation. 

Description: \{caption\} 

Output:

\end{tcolorbox}

\section{Detailed Zero-Shot Results}

\begin{table*}[t]
  \centering
      \begin{tabular}{l|cc|ccc}
        \hline
        Dataset & ODinW13 & ODinW35 & G-DINO-T & MM-GDINO-T & LLMDet \\
        \hline
        AerialMaritimeDrone large & \checkmark & \checkmark & 0.173 & 0.155 & 0.153 \\
        AerialMaritimeDrone tiled  &  & \checkmark & 0.206 & 0.201 & 0.174 \\
        AmericanSignLanguageLetters &  & \checkmark & 0.002 & 0.007 & 0.016 \\
        Aquarium & \checkmark & \checkmark & 0.195 & 0.281 & 0.268 \\
        BCCD &  & \checkmark & 0.161 & 0.078 & 0.149 \\
        boggleBoards &  & \checkmark & 0.000 & 0.002 & 0.001 \\
        brackishUnderwater &  & \checkmark & 0.021 & 0.024 & 0.026 \\
        ChessPieces &  & \checkmark & 0.000 & 0.000 & 0.000 \\
        CottontailRabbits & \checkmark & \checkmark & 0.806 & 0.788 & 0.797 \\
        dice &  & \checkmark & 0.004 & 0.001 & 0.004 \\
        DroneControl &  & \checkmark & 0.042 & 0.073 & 0.070 \\
        EgoHands generic & \checkmark & \checkmark & 0.608 & 0.518 & 0.518 \\
        EgoHands specific &  & \checkmark & 0.002 & 0.003 & 0.010 \\
        HardHatWorkers &  & \checkmark & 0.046 & 0.109 & 0.178 \\
        MaskWearing &  & \checkmark & 0.004 & 0.009 & 0.004 \\
        MountainDewCommercial &  & \checkmark & 0.430 & 0.433 & 0.518 \\
        NorthAmericaMushrooms & \checkmark & \checkmark & 0.471 & 0.747 & 0.749 \\
        openPoetryVision &  & \checkmark & 0.000 & 0.000 & 0.003 \\
        OxfordPets by breed &  & \checkmark & 0.003 & 0.004 & 0.006 \\
        OxfordPets by species & & \checkmark & 0.011 & 0.016 & 0.024 \\
        PKLot &  & \checkmark & 0.001 & 0.007 & 0.034 \\
        Packages & \checkmark & \checkmark & 0.695 & 0.706 & 0.717 \\
        PascalVOC & \checkmark & \checkmark & 0.563 & 0.566 & 0.584 \\
        pistols & \checkmark & \checkmark & 0.726 & 0.726 & 0.720 \\
        plantdoc &  & \checkmark & 0.005 & 0.011 & 0.005 \\
        pothole & \checkmark & \checkmark & 0.215 & 0.164 & 0.175 \\
        Raccoons & \checkmark & \checkmark & 0.549 & 0.533 & 0.519 \\
        selfdrivingCar &  & \checkmark & 0.089 & 0.082 & 0.083 \\
        ShellfishOpenImages & \checkmark & \checkmark & 0.393 & 0.489 & 0.429 \\
        ThermalCheetah &  & \checkmark & 0.087 & 0.045 & 0.132 \\
        thermalDogsAndPeople & \checkmark & \checkmark & 0.657 & 0.548 & 0.546 \\
        UnoCards &  & \checkmark & 0.006 & 0.005 & 0.010 \\
        VehiclesOpenImages & \checkmark & \checkmark & 0.613 & 0.610 & 0.597 \\
        WildfireSmoke &  & \checkmark & 0.134 & 0.129 & 0.093 \\
        websiteScreenshots &  & \checkmark & 0.012 & 0.016 & 0.013 \\
        \hline
        ODinW13 Average &  &  & 0.514 & 0.525 & 0.521 \\
        ODinW35 Average &  &  & 0.227 & 0.231 & 0.238 \\
        \hline
      \end{tabular}
  \caption{Detailed zero-shot results on ODinW35~\cite{li2022elevater}.}
  \label{tab:detailed_odinw}
\end{table*}

\begin{table*}[t]
  \centering
      \begin{tabular}{l|cccccc|c}
        \hline
        Model & Cartoon & Handmake & Painting & Sketch & Tattoo & Weather & Average \\
        \hline
        Grounding-DINO~\cite{grounding_dino} & 40.2 & 30.2 & 43.1 & 37.6 & 29.8 & 44.8 & 37.6 \\
        MM-GDINO~\cite{mm_GDINO} & 35.0 & 26.6 & 41.7 & 32.2 & 23.9 & 44.8 & 34.0 \\
        LLMDet & 37.7 & 30.7 & 42.8 & 32.6 & 27.5 & 45.3 & 36.1 \\
        \hline
      \end{tabular}
  \caption{Detailed zero-shot results on COCO-O~\cite{mao2023coco}.}
  \label{tab:detailed_coco_o}
\end{table*}

\noindent{\textbf{Detailed zero-shot results on ODinW35.}} \Cref{tab:detailed_odinw} lists the detailed performance of Grounding-DINO-T~\cite{grounding_dino}, MM-GDINO-T~\cite{mm_GDINO}, and our LLMDet on each dataset in ODinW35~\cite{li2022elevater}. The selected datasets in ODinW13 are also marked out.

\vspace{0.5em}\noindent{\textbf{Detailed zero-shot results on COCO-O.}} COCO-O~\cite{mao2023coco} is a dataset sharing the same 80 classes as COCO but in different domains including cartoon, handmake, painting, sketch, tattoo, and weather. Detailed performance on each domain is listed in \Cref{tab:detailed_coco_o}.

\section{Visualization}
\subsection{Visualizations of the Image-Level Captions in GroundingCap-1M}

\begin{figure*}[t]
  \centering
  \vspace{-0.6em}
  \includegraphics[width=0.9\linewidth]{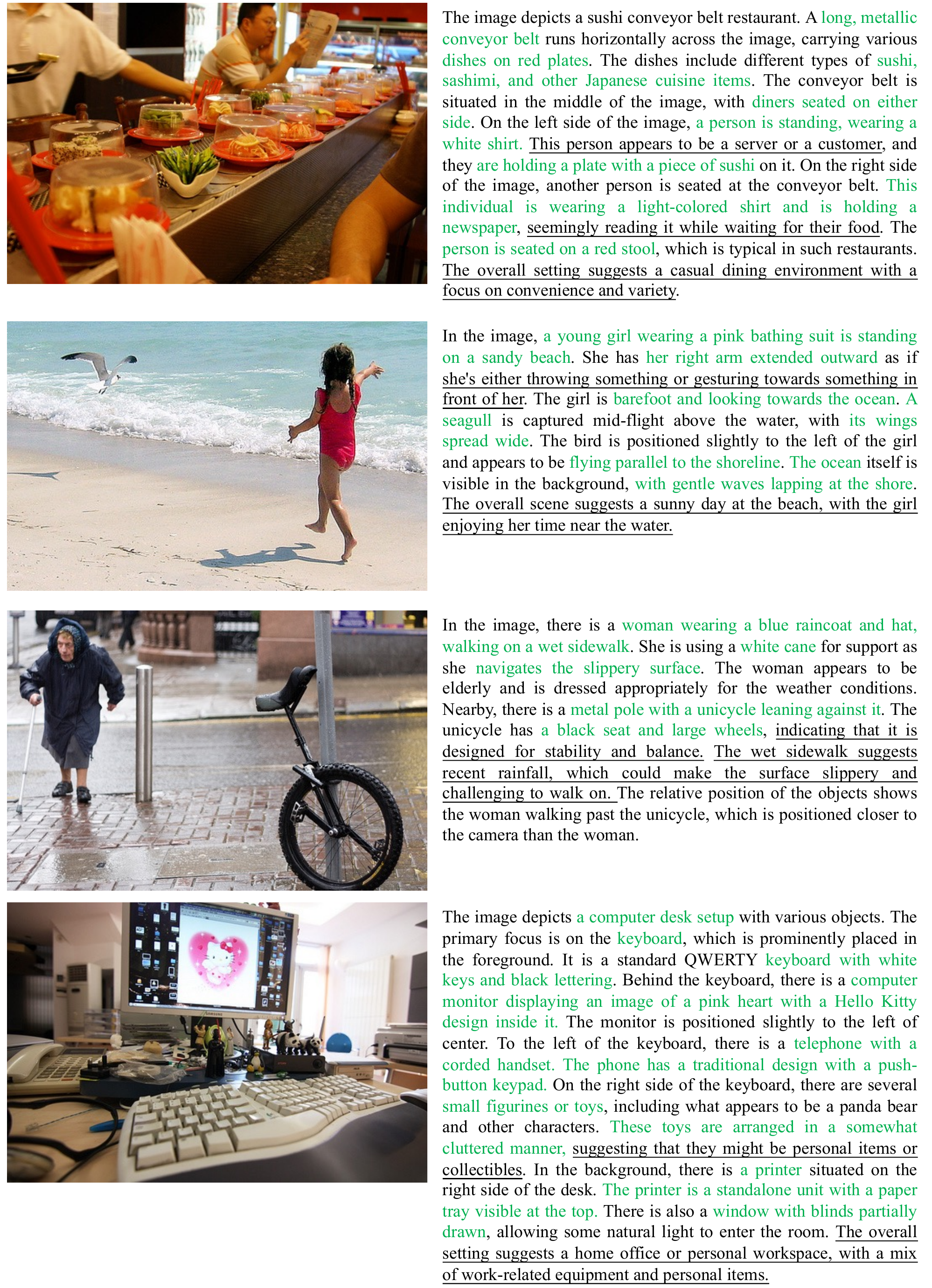}
  \caption{Visualizations of the image-level captions in GroundingCap-1M, which are rich in detail. The great details are marked in green color. But the captions still contain some imaginary contents, which are also highlighted by underlines.}
  \label{fig:caption_visualization}
\end{figure*}

In this work, we collect a new GroundingCap-1M dataset which equips a standard grounding dataset with detailed image-level captions. The captions should contain as many details as possible, including object types, textures, colors, parts of the objects, object actions, precise object locations, and texts. And the captions should not contain imaginary contents. \Cref{fig:caption_visualization} visualizes some examples in GroundingCap-1M. The captions shown depict the main entities in the pictures with great detail (demonstrated in green color) but also with some imaginary contents inevitably (also highlighted by underlines). The imaginary contents always start with speculative words, like ``seemingly'', ``indicating'', and ``suggesting''. We just find these pre-defined speculative words and delete the sub-sentences including them in an online manner.

\subsection{Visualizations of the Captions Generated by LLMDet}

\begin{figure*}[t]
  \centering
  \vspace{-0.6em}
  \includegraphics[width=0.9\linewidth]{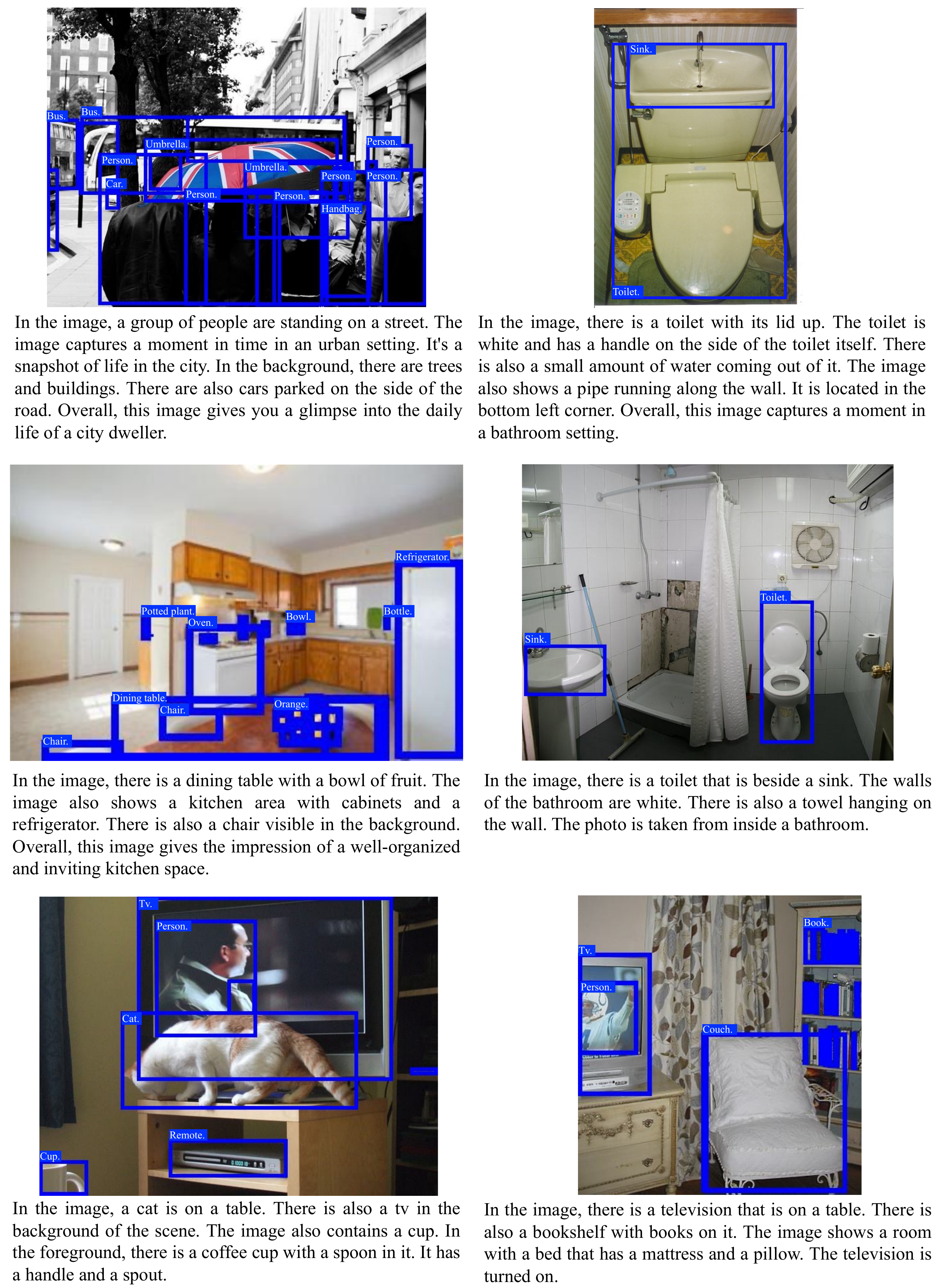}
  \caption{Visualizations of the image-level and region-level generated captions from the LLM co-trained with LLMDet. Image-level captions are placed under the corresponding images and region-level captions are placed beside the bounding boxes. Only object queries with scores higher than 0.3 are visualized in the images.}
  \label{fig:caption_generation}
\end{figure*}

In \Cref{fig:caption_generation}, we visualize some examples of the generated image-level and region-level captions from the LLM co-trained with LLMDet. Images are selected from the COCO validation set. The LLM can generate precise class names for the objects in COCO (as we use the class names in COCO as the grounding text for deep fusion, only objects in COCO are detected out for caption generation). But we find that the image-level captions are relatively coarse-grained compared with the ones in GroundingCap-1M. We suppose the reason is that our region-level data is far more than the image-level data (one image has multiple regions) and the region-level data is overly simplistic.

\end{appendices}


\end{document}